%% file: OVNNI.tex
\documentclass[runningheads]{llncs}
\usepackage{subfig}
\usepackage{graphicx}
\usepackage{comment}
\usepackage{amsmath,amssymb} 
\usepackage{color}

\usepackage{microtype}
\usepackage{booktabs}       
\usepackage{multirow}
\usepackage{wrapfig}

\definecolor{orange}{rgb}{0.99,0.29,0.07}

\newcommand\Gianni{\textcolor{black}}

\newcommand{\ubold}[1]{\fontseries{b}\selectfont#1}

\newcommand{\vomega}{\boldsymbol{\mathbf{\omega}}}

\newcommand{\etal}{\textit{et al}.}

\newcommand{\eg}{\textit{e}.\textit{g}.}

\begin{document}
\pagestyle{headings}
\mainmatter
\def\ECCVSubNumber{2692}  

\title{One Versus all for deep Neural Network Incertitude (OVNNI) quantification} 

\titlerunning{One Versus all for deep Neural Network Incertitude (OVNNI) quantification}
%
\author{Gianni Franchi\inst{1,2} \and
Andrei Bursuc\inst{3} \and
Emanuel Aldea\inst{2} \and
S\'{e}verine Dubuisson\inst{4}   \and
Isabelle Bloch\inst{5} 
}

\authorrunning{Franchi et al.}
%
\institute{ENSTA  Paris, Institut polytechnique de Paris \and
SATIE, Universit\'{e}  Paris-Sud, Universit\'{e} Paris-Saclay \and
valeo.ai \and
CNRS, LIS, Aix Marseille University \and
LTCI, T\'{e}l\'{e}com Paris, Institut polytechnique de Paris}
\maketitle

\begin{abstract}
Deep neural networks (DNNs) are powerful learning models yet their results are not always reliable. This is due to the fact that modern DNNs are usually uncalibrated and we cannot characterize their epistemic uncertainty.  
In this work, we propose a new technique to quantify the epistemic uncertainty of data easily. This method consists in mixing the predictions of an ensemble of DNNs trained to classify One class vs All the other classes (OVA) with predictions from a
standard DNN trained to perform  All vs All (AVA) classification. 
On the one hand, the adjustment provided by the AVA DNN to the score of the base classifiers allows for a more fine-grained inter-class separation. On the other hand, the two types of classifiers enforce mutually their detection of out-of-distribution (OOD) samples, circumventing entirely the requirement of using such samples during training.
Our method achieves state of the art performance in quantifying OOD data across multiple datasets and architectures while requiring little hyper-parameter tuning.

\keywords{Uncertainty estimation, DNN ensembles, one vs all classification, all vs all classification.}
\end{abstract}

\section{Introduction}

Deep neural networks (DNNs) have reached state-of-the-art performance on machine learning \cite{lecun2015deep,goodfellow2014generative}, and computer vision tasks \cite{ren2015faster,krizhevsky2012imagenet}. The significant progress has been leading to their adoption in a wide range of decision making systems, including safety critical ones. Yet, one of the main weaknesses of these techniques appears to be the fact that they tend to be overconfident \cite{guo2017calibration} in their decisions. This issue is difficult to tackle, as the high inner complexity of DNNs results in a poor output explainability. 

\begin{figure}[!thb]
     \centering
     \subfloat[]{\includegraphics[width=0.30\linewidth]{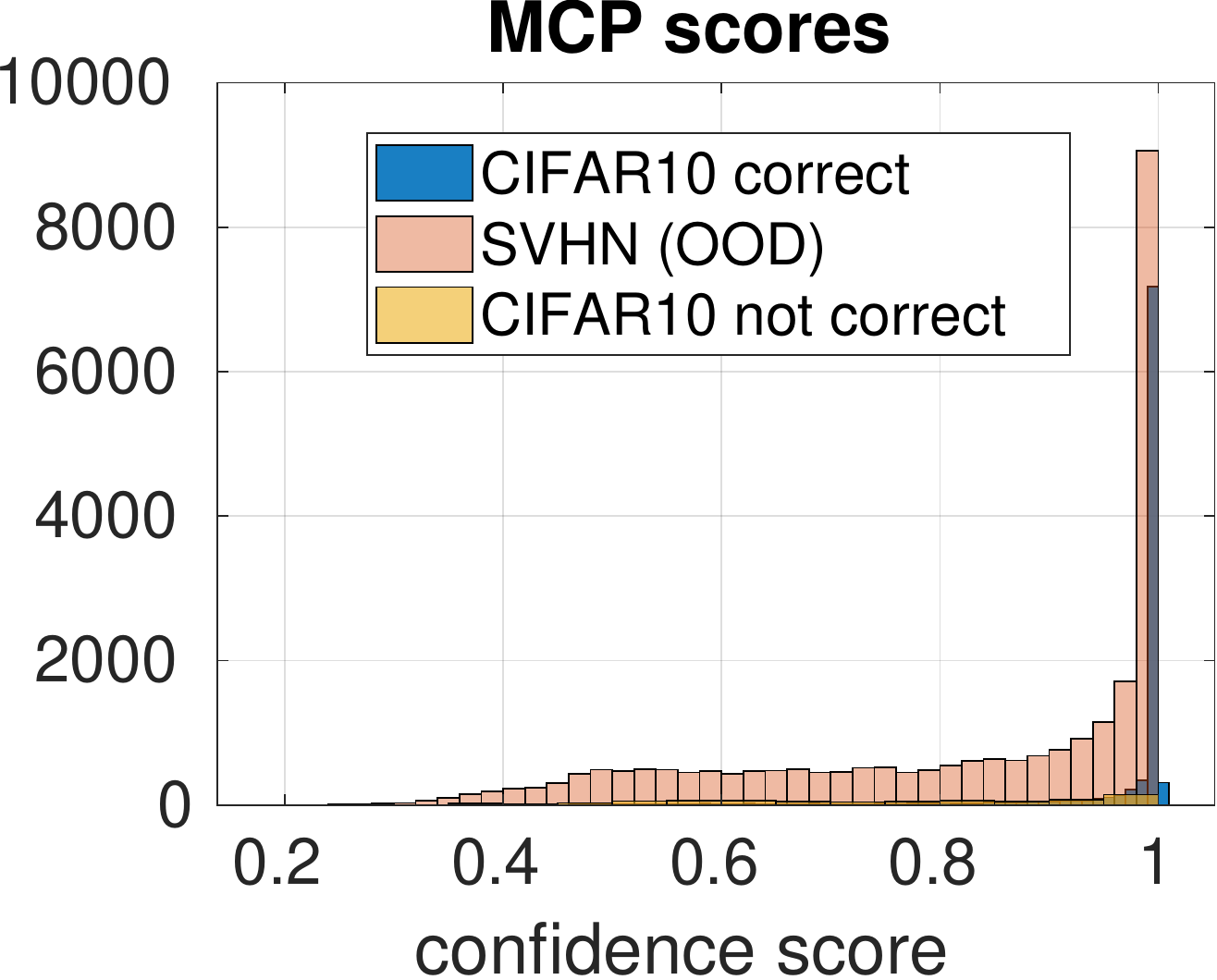}\label{histo_DE}}
    \subfloat[]{\includegraphics[width=0.30\linewidth]{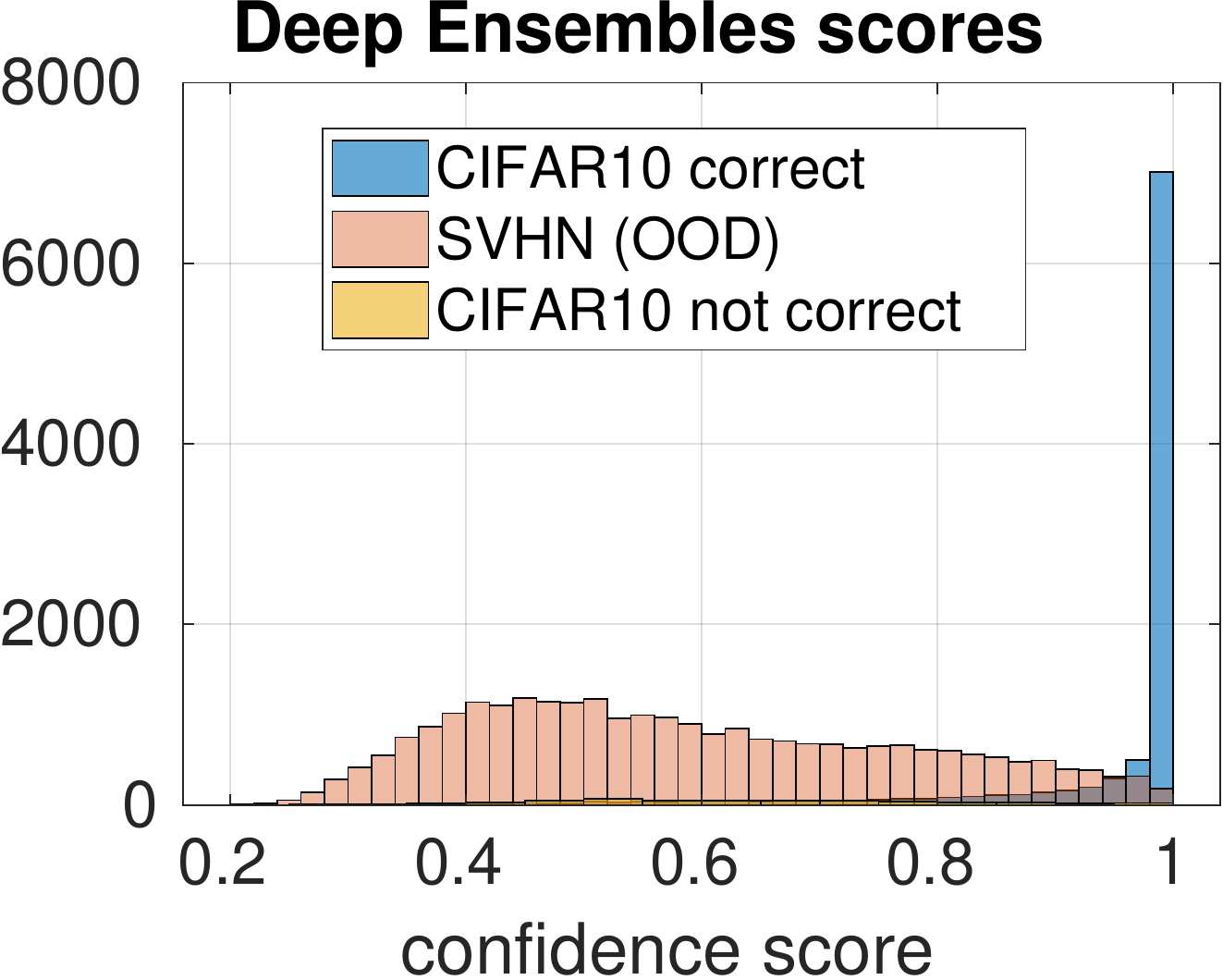}\label{histo_DE}}
    \subfloat[]{\includegraphics[width=0.30\linewidth]{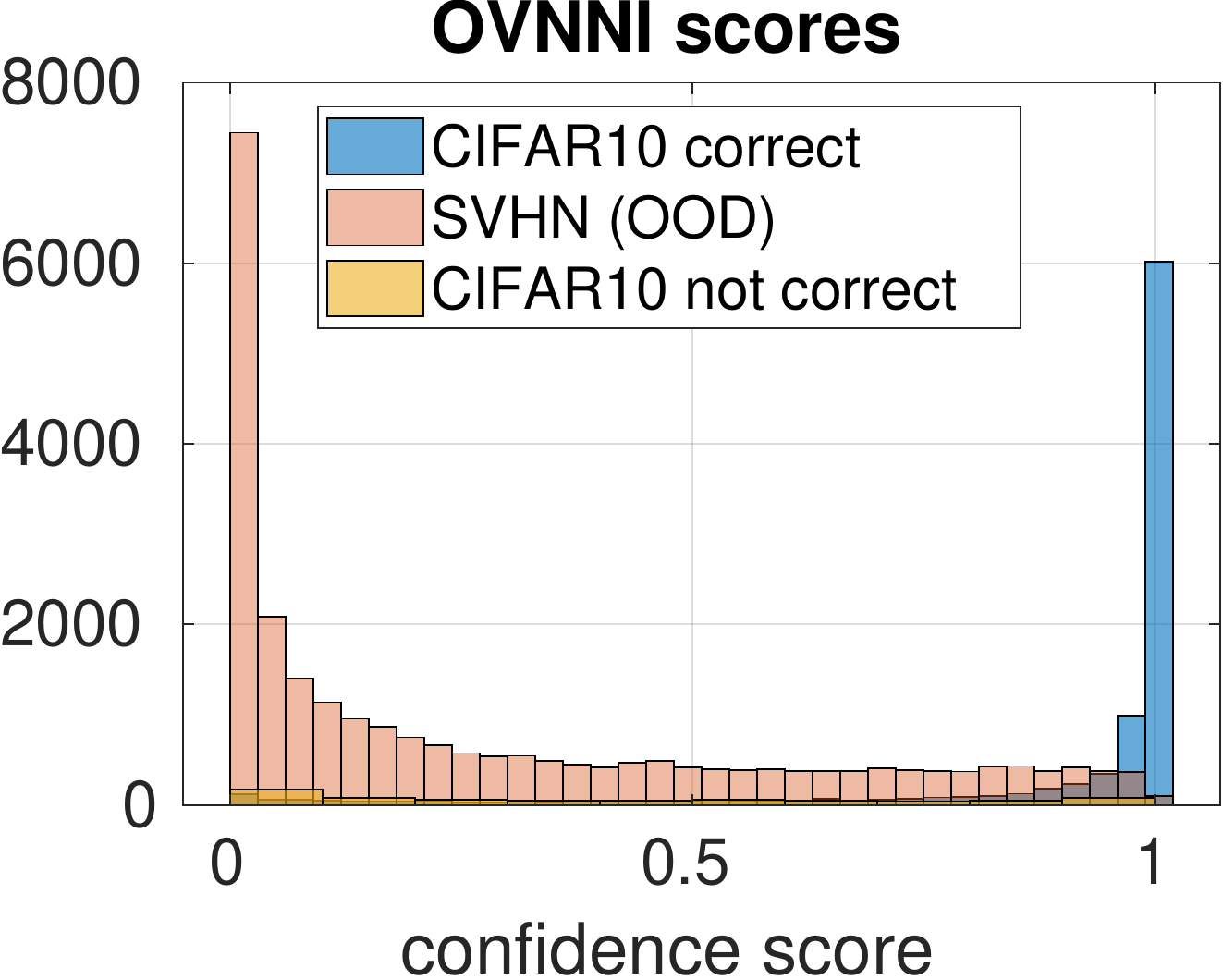}\label{histo_OVA}}
    \caption{Distribution of classifications scores. Here (a), (b) and (c) represent the histograms of confidence scores of Resnet50 \cite{he2016deep} trained on the CIFAR10~\cite{krizhevsky2009learning} training set and tested on SVHN \cite{Netzer2011} and CIFAR10 testing set, using Maximum Class Probability (MCP) \cite{hendrycks2016baseline}, Deep Ensembles~\cite{lakshminarayanan2017simple}, and OVNNI, respectively. We can see that our proposed algorithm OVNNI outperforms Deep Ensembles (state of the art)  and MCP (baseline) on detecting OOD data, since it brings more OOD data to a low score.}
     \label{fig:histos}
         \end{figure}

In order to address this crucial issue, we propose to rely on a finer quantification of the uncertainty of DNN. In contrast to most Bayesian DNN techniques \cite{blundell2015weight,kendall2017uncertainties,gal2016dropout,maddox2019simple,franchi2019tradi}, or to frequentist techniques such as Deep Ensembles~\cite{lakshminarayanan2017simple}, our approach relies on One vs All (OVA) training. In the statistical learning community, ensembles of OVA or One vs One (OVO) base classifiers for multi-class prediction have been particularly popular in association with Support Vector Machines (SVM), due to SVM being 
essentially a binary classifier, and to the simplicity of the aggregation rules supported by fundamental theoretical results~\cite{kittler1998combining,kuncheva2002theoretical,tumer1996error}. The most popular rule in case of OVA ensembles, \textit{winner-takes-all} (WTA), assigns the testing sample to the class for which the membership score is the highest. For a binary output, the WTA rule creates in the input space multiple unclassifiable regions, for which the class assignment is not unique, and the standard solution is to rely on continuous membership scores. In contrast to SVM-based learning, nowadays the OVA approach has been mostly discarded when training deep classifiers, in favor of All vs All (AVA) learning. 

In this paper, we propose to use OVA learning in order to 
improve the quantification of the epistemic uncertainty of the DNN. The underlying idea of our approach is that the score of a base classifier should be adjusted by a factor which approximates its local reliability in the input space from which the test sample originated. Initially for SVM learning, the reliability has been linked to the average value of the \textit{local} objective function~\cite{liu2005one}, which is approximated using the closest training samples belonging to the respective class. In our algorithm, we propose to adjust the OVA scores by the score provided by an AVA DNN which will play thus the role of approximating the local class-specific objective function. This strategy allows for a particularly effective detection of out-of-distribution (OOD) samples in the testing data, as we can discriminate between samples belonging to unclassifiable regions equally close to \textit{some} classifiable regions, and samples belonging to unclassifiable regions far from \textit{all} classifiable regions. 

Fig.~\ref{fig:histos} presents the distribution of the scores provided by the baseline, Deep Ensembles (the current state of the art) and our method, respectively. The baseline is the single AVA classifier, for which the class assignment is performed based on the Maximum Class Probability (MCP). The baseline is unable to discriminate among in- and out-of-distribution samples, illustrated in blue/yellow and orange in the histograms, respectively. Deep Ensembles lowers the OOD scores, but the in-distribution membership is still overestimated. Finally, OVNNI successfully assigns low scores to the OOD samples, while keeping at the same time the in-distribution scores high.

Our main contributions are the following. We propose an efficient non-Bayesian technique for uncertainty quantification in OOD data classification, that reaches state of the art results on calibration and on OOD data detection on a variety of datasets, and on all typical metrics. Secondly, we perform an extensive study in which we compare, for different applications, with the most significant approaches tackling uncertainty estimation, including Deep Ensembles. Lastly, our conclusions are in line with those of other works which defend the interest of One versus All classifier aggregation~\cite{rifkin2004defense}, and our results rehabilitate this approach in the novel context of uncertainty estimation for DNN.

\input{related.tex}

 \section{One Versus all for deep Neural Network Incertitude (OVNNI)}
 
This section focuses first on the necessary details on the traditional AVA training of a DNN. Then we describe our approach based on additional OVA training.

\subsection{Notations}

\begin{itemize}

\item The training/testing sets {are} denoted respectively {by} $\mathcal{D}_l=(x_i,y_i)_{i=1}^{n_l}$,   $\mathcal{D}_{\tau}=(x_i,y_i)_{i=1}^{n_{\tau}}$, where $x_i$ and $y_i$ with $i \in \{1... n_l\}$ or  $i \in \{1... n_{\tau}\}$ 
represent respectively the observed sample and the corresponding label, with $n_l$ and $n_{\tau}$ {the size of the training and testing sets.} {$x_i$ are input vectors and $y_i\in \{0,\ldots,n_{\mbox{\tiny label}}\}$ are class labels.} {Unless otherwise specified, $x_i$ and $y_i$, $i \in [1,n_l]$, will refer to training data.}

\item $X$ is the random variable associated with observed samples and $Y$ the one associated with classes.


\item 
{The DNN is a} function $f$ of the observed data $x_i$ with $i \in [1,n_l]$ or  $i \in [1,n_{\tau}]$ and vector
$\vomega$ that contains the trainable weights.
{We call} $f_{\vomega}(x_i)$ the output of the DNN associated with the weights $\vomega$ on the data  $x_i$.

\item $\mathcal{L}(\vomega,y_i)$ is the loss function used to measure the dissimilarity between the output $f_{\vomega}(x_i)$ of the DNN  and the expected output $y_i$. Different loss functions can be considered according to the type of task. Here we will focus on the cross entropy that will be introduced in the next section.

\end{itemize}

 \subsection{All Versus All training of Deep Neural Networks}

{For image classification, the goal of a DNN is}
to map the input data to a probabilistic prediction that we denote $P(Y=y^\ast \mid X=x_i, \vomega)$ with $y^\ast$ {a class label}. 
During training, an optimization algorithm will improve the weights  $\vomega$ in order to fit as much as possible the output to the ground truth vector of class labels. 
The loss is expected to measure the similarity between $f_{\vomega}(x_i)$ and $y_i$. Classically we use Cross entropy defined on a batch  $B$ of size $N \in \mathbb{N}$ by: 
   \begin{eqnarray}\label{Cross_entropy}
\mathcal{L}(\vomega(t),B)  = - \frac{1}{N}\sum_{i=1}^N \mathcal{L}(\vomega(t),y_i)=- \frac{1}{N}\sum_{i=1}^N \log (P(Y=y_i \mid X=x_i, \vomega))
 \end{eqnarray}

The minimization of this loss function is usually based on gradient methods. Computing the optimal value of each parameter involves a bin-to-bin measure of similarity, which may lead to overfitting issues.

A solution might be to use One Versus All training.

  \subsection{From One Versus All (OVA)  to OVNNI}
  
The current state of the art on uncertainty estimation  is Deep Ensembles \cite{lakshminarayanan2017simple}. This technique relies on ensembling multiple DNN models trained in parallel in order to optimize the same loss. In contrast  to random forests \cite{breiman2001random}, or Bagging\cite{breiman1996bagging} the diversity arises from the fact that different embodiments of the same model will converge towards different local optima during training. Conversely, in our approach the diversity is provided by the one-versus-all (OVA) models constructed using different labelings of the training set.

 The OVA strategy is conceptually simple, since at its core it involves training a binary classification DNN.  One classifier is trained for each class, and prediction is then performed by running the obtained binary classifiers  on the testing sample and choosing the prediction with the highest confidence score. Yet, the multiple classifiers involved will learn multiple probabilistic predictions, denoted by $P(Y_j=1 \mid X=x_i, \vomega^j)$ with $Y_j$ a binary random variable for each class $j$. We add a super script on $\vomega^j$, to inform that 1) weights are different from the ones trained to perform the AVA classification that we denote $\vomega$, and 2) they are also different from the weights of other classes different of $j$. 
 
\begin{figure}[t!]
\renewcommand{\captionfont}{\small}
\begin{center}

\includegraphics[width=0.95\columnwidth]{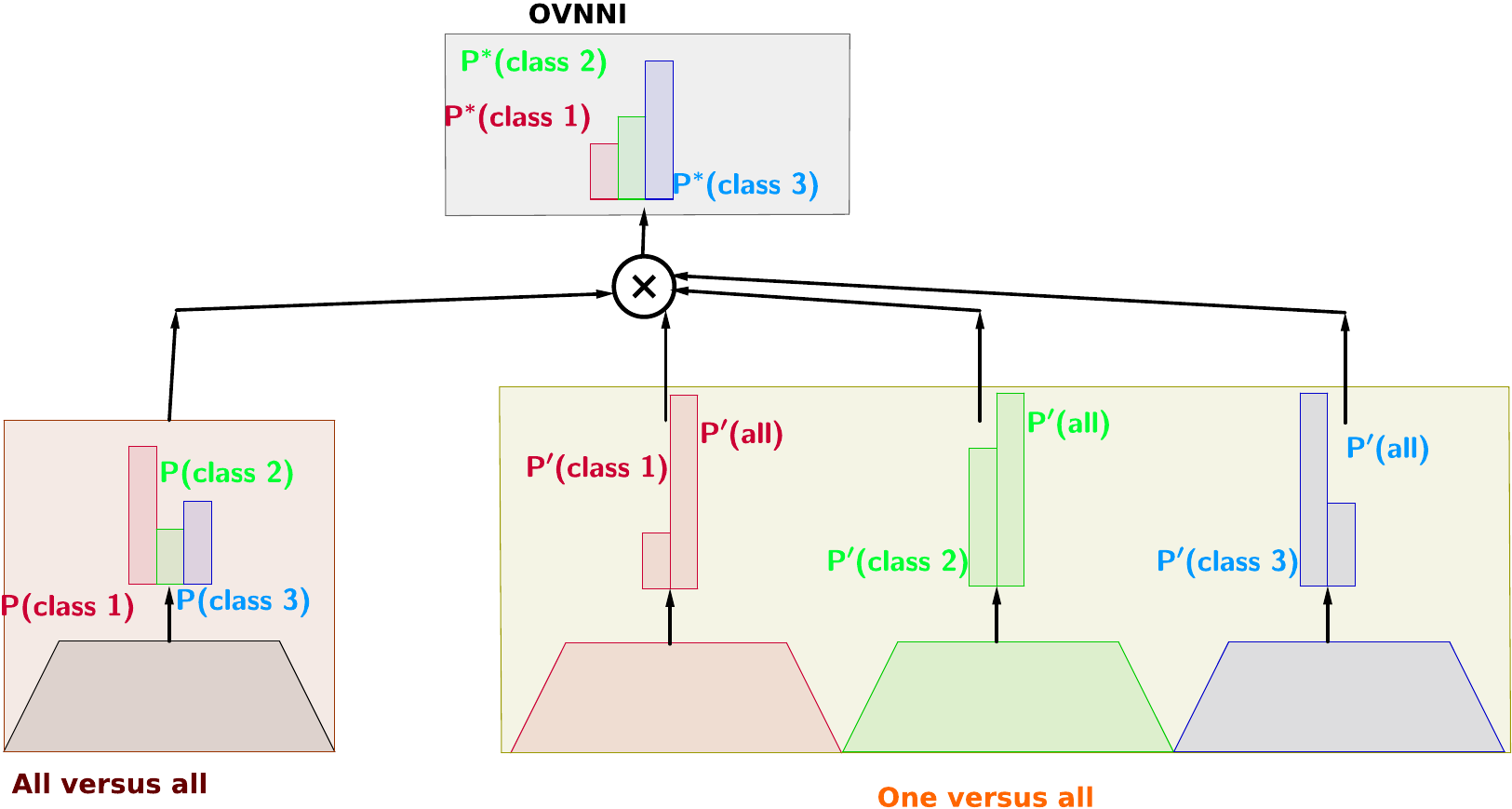}

\end{center}
\vspace{-0.5cm}
\caption{ From AVA and OVA to OVNNI process in the case we deal with a database composed of just three classes. }\label{fig:process}
\end{figure}

 By training one class versus all the other classes, the DNN learns in some sense the out of distribution classes, however with the significant advantage of not relying on explicitly provided OOD data, in contrast to other strategies \cite{papadopoulos2019outlier,masana2018metric}. In addition to the OVA base classifiers, we also perform an All versus all training that we aggregate with the probabilities of the OVA models in the following way as shown in Figure \ref{fig:process}. 

Let us denote by $Y$ the discrete random variable, that is taking its value in the list of all classes, and let us denote by $Y_j$ a binary random variable that takes values $0$ or $1$, with $Y_j=1$ meaning that the data belongs to class $j$. Hence the OVA  DNN of the class $j$ provides $P(Y_j=1 \mid X=x_i, \vomega^j)$, while the AVA DNN provides $P(Y=j \mid X=x_i, \vomega)$ for all $j$ in $[1,n_{\mbox{\tiny label}}]$. We consider that the final \Gianni{confidence score} 
for a data $x_i$ to belong to class $j$ is:
   \begin{eqnarray}\label{OVA_AVA}%
p_j(x_i)=P(Y_j=1 \mid X=x_i, \vomega^{j}) \times P(Y=j \mid X=x_i, \vomega)
 \end{eqnarray}
\Gianni{This score is high if AVA and OVA are confident and low in the other case, multiplying OVA and AVA  scores also helps to increase the accuracy since AVA has lower accuracy than OVA.}

\subsection{Uncertainty with OVNNI}

We consider that a measure of confidence must satisfy the following properties: (1) be bounded, (2) exhibit low values for OOD data,
(3) have a confidence value that  aligned to the accuracy of the algorithm, (4) get more confident if  additional training samples are provided.
The first point assures that we know what is the maximum and minimum of confidence. The second point is to {ensure to} 
detect OOD data, which is crucial since {it provides} information on the reliability of the DNN on one data. The third point is linked {to} the calibration \cite{guo2017calibration}, which is crucial to rely on the model predictions.
The last point {concerns the fact that we want to reduce the uncertainty when increasing the dataset.}

We use as a measure of confidence  for OVNNI the probability $\displaystyle \max_{j\in [1,n_{\mbox{\tiny label}}]}\{p_j(x_i)\}$. This measure is bounded by 0 and 1. In the experimental section, we show that it has a state of the art calibration and OOD results.

\subsection{Visualizing OVA and AVA embedding}

In this subsection, we perform two experiments to determine the behavior of the representations learned by the DNN with the different techniques. For both experiments we train a simple DNN  composed of 3 hidden layers followed by a batch normalization on MNIST dataset \cite{lecun1998gradient}.

\begin{figure}[!thb]
     \centering
         \subfloat[][]{\includegraphics[width=0.33\linewidth]{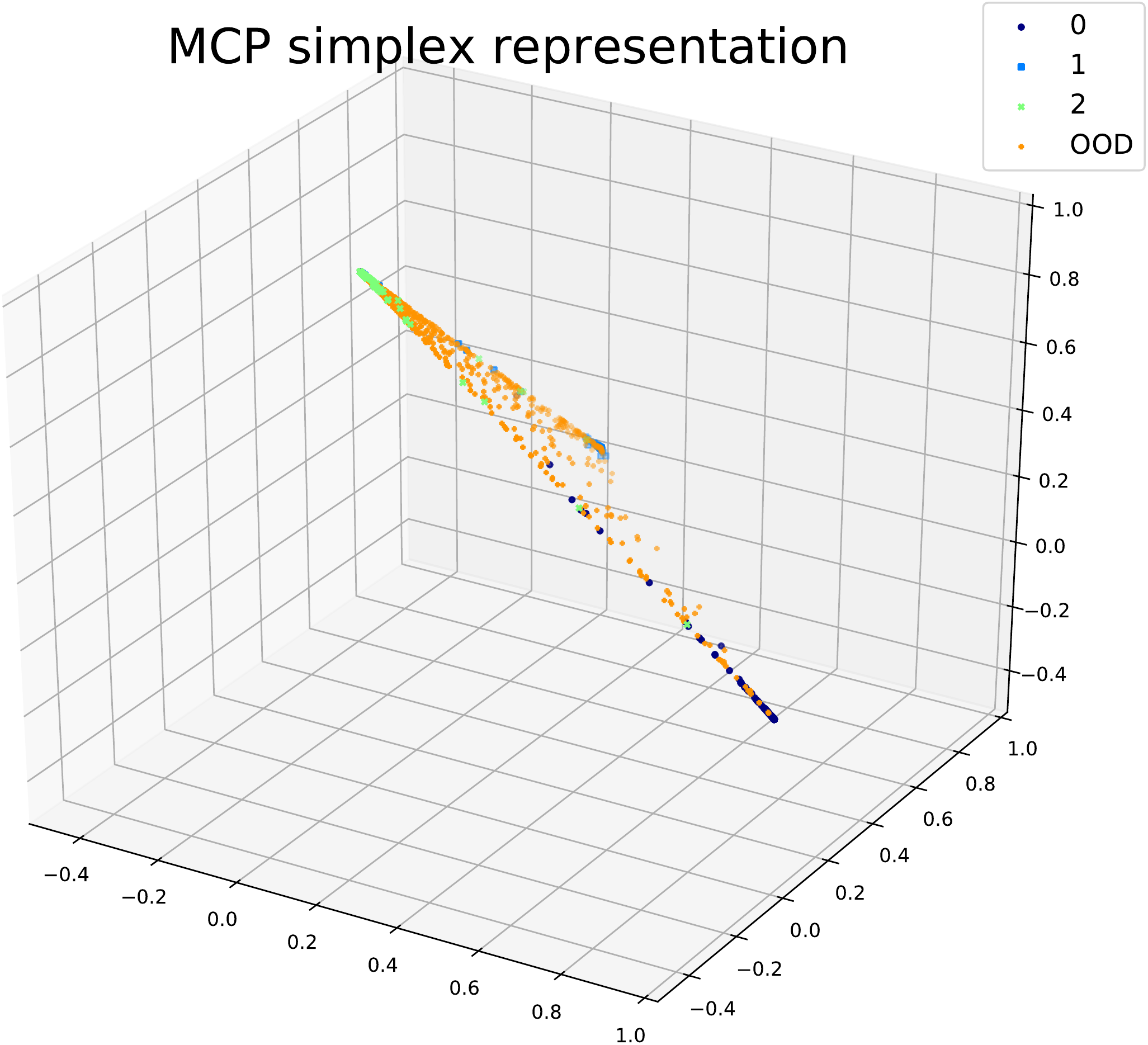}}
         \subfloat[][]{\includegraphics[width=0.33\linewidth]{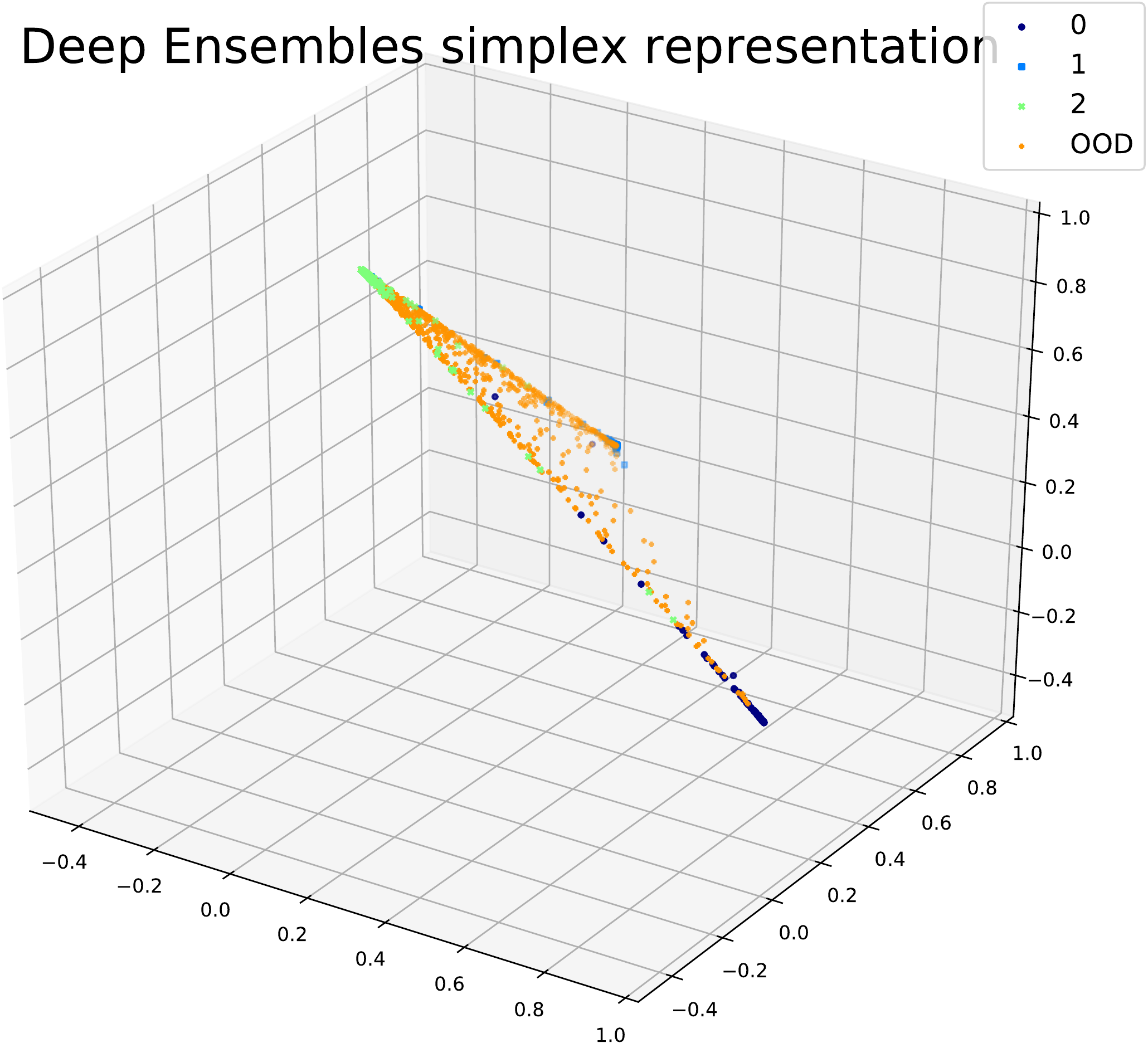}}
         \subfloat[][]{\includegraphics[width=0.33\linewidth]{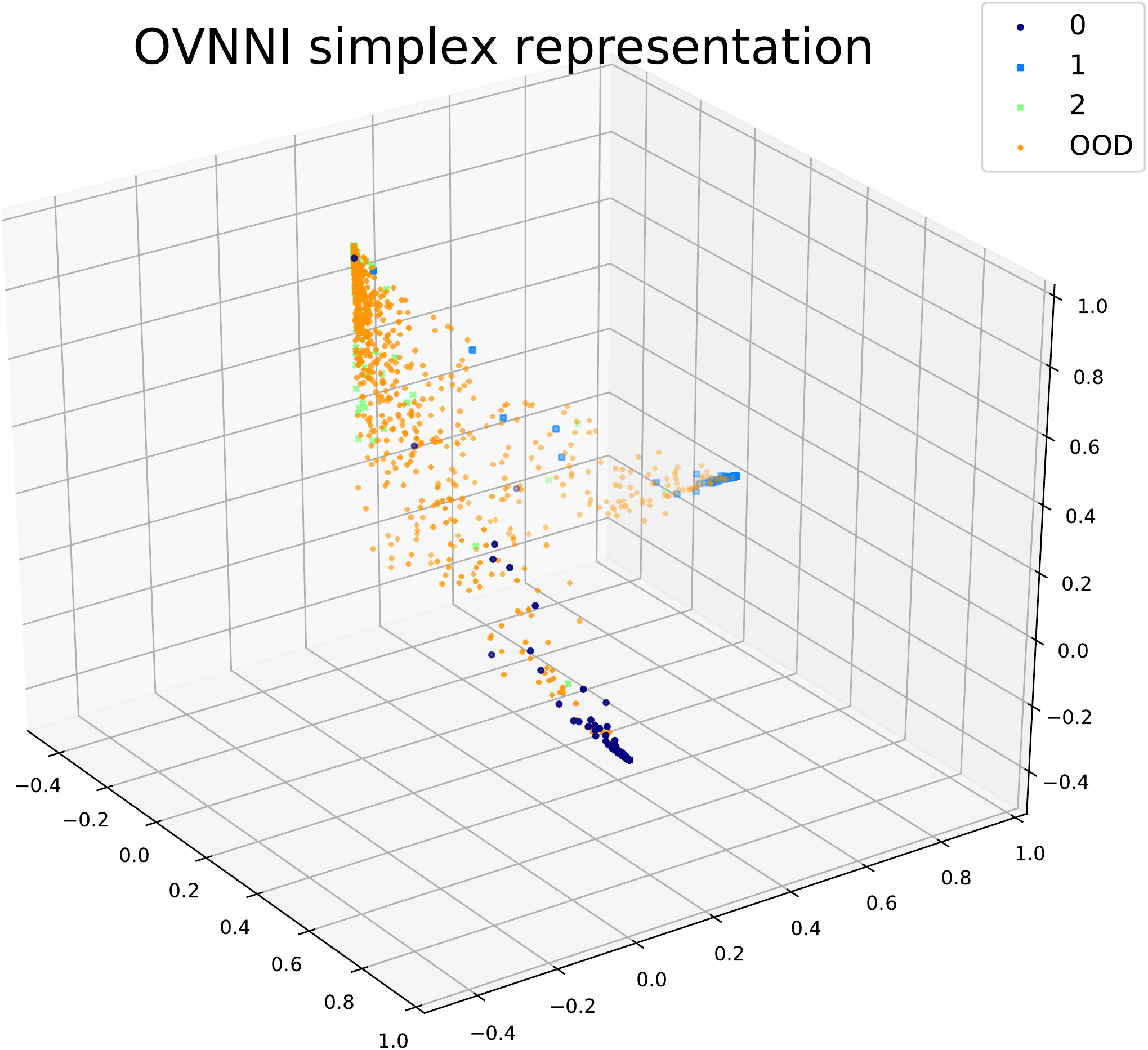}}
   \caption{Results on MNIST - 3 classes experiments. We represent in these figures the softmax prediction outputs obtained by the baselines (a) MCP, (b) Deep Ensemble, and (c) by OVNNI, respectively.
}\label{fig:simplex}
         \end{figure}

In the first experiment, we have considered as training data only the images with the digits '0','1' and '2' images (the 3 first classes).  Then we perform inference on the official test set composed of images with these classes and the OOD images which are composed of other classes. We represent in Figure \ref{fig:simplex} the softmax of a classical AVA training, a deep ensemble training and the OVNNI training. One can see that in contrast to other techniques, OVNNI results do not necessarily belong to the 2-dimensional simplex. In addition, OVNNI brings the OOD data far away from the simplex vertices which highlights its potential to detect OOD data.

\begin{figure}[!thb]
     \centering
         \subfloat[][MCP PCA]{\includegraphics[width=0.25\linewidth]{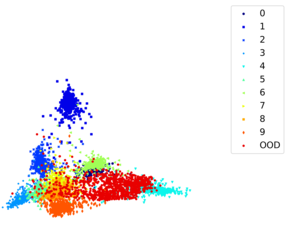}}
         \subfloat[][OVNNI PCA]{\includegraphics[width=0.25\linewidth]{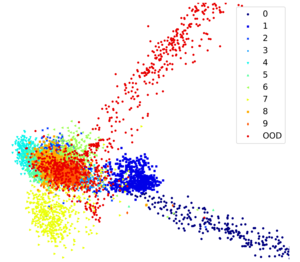}}
         \subfloat[][MCP t-SNE]{\includegraphics[width=0.25\linewidth]{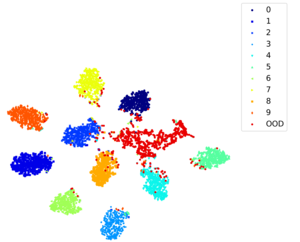}}
        \subfloat[][OVNNI t-SNE]{\includegraphics[width=0.25\linewidth]{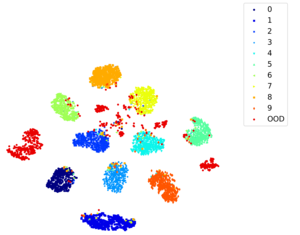}}
   \caption{Results of the MNIST / NotMNIST experiment. We represent the projection on a 2D space of the feature space of the baseline MCP in figures (a) and (c), and of OVNNI in figures(b) and (d). We use PCA \cite{wold1987principal} and  t-SNE \cite{maaten2008visualizing} as dimensionality reduction algorithm.
}\label{fig:feature_space}
         \end{figure}

{In the second experiment},  we performed a classical AVA training, and we also performed the OVA training. Hence for the OVA training, we have 10 DNNs (since the dataset has 10 classes which are the 10 digits).
The OOD class is composed of images of the NotMNIST dataset~\cite{NotMnist}.
Hence, we apply the DNNs on this test dataset and on the AVA case, we collect for each data the feature space of the DNN just before the classification of each data. {In the OVA case, we collect the same feature space but for the DNN of the predicted class.}
We reduce the dimension of each of these feature spaces using T-SNE \cite{maaten2008visualizing} and Principal Component Analysis (PCA) \cite{wold1987principal} and we plot the results in Figure \ref{fig:feature_space}.
We can see that in the AVA case the OOD data are in the center {of Figure \ref{fig:feature_space} mixed with the other classes} and in the OVA case they are closer to the border whatever the dimensionality reduction algorithm we use. This is crucial because it shows that OVA learns a more interesting descriptor than OVA.


\section{Experiments}
 We continue by illustrating the performance of OVNNI for detecting OOD data by conducting five experiments. In the rest of this section we will describe the experimental protocol, followed by the five experiments.

\subsection{Experimental protocol}

The detection of OOD data can be done either by techniques that measure the uncertainty, or by techniques that detect OOD data.
{We first have compared our OVNNI to three other uncertainty estimation techniques: MC Dropout \cite{gal2016dropout}, Deep Ensembles \cite{lakshminarayanan2017simple}, and TRADI \cite{franchi2019tradi}. }
The major interest of these techniques comes from the fact that, since they estimate uncertainty, they also estimate the epistemic uncertainty and therefore the OOD data. We also {have} compared {our approach} to two {other} techniques: ODIN \cite{liang2017enhancing} and ConfidNET \cite{corbiere2019addressing} {and} serve as references in unsupervised techniques for detecting OOD data. As a baseline algorithm, we use the maximum class probability (MCP)  with AVA trained DNN.  We denote this approach as MCP. 
\Gianni{As an additional baseline we consider one-class Support Vector Machine\cite{moya1996network,oliveri2017class}, a classic method for outlier detection. We train it on AVA logits.}

Note that we have not compared our OVNNI to techniques trained to \textit{learn} OOD such as  \cite{papadopoulos2019outlier,masana2018metric}, since in these case the OOD data are in  the training set, making this technique able to detect just with trained OOD data. 
\Gianni{To balance OVA training 
which typically has more samples available for the ''All'' class,
we use weighted cross-entropy to train for each class, with weights for a given class based on $1 -\tau_{\mbox{class}} $, where $\tau_{class}$ is the proportion of data samples of this class in the training set. In addition, for a fair comparison in all experiments we use the same number of models for ensemble and Bayesian methods.}
{We conducted several experiments in two target applications:}
 image classification (2 experiments) and semantic pixel segmentation (3 experiments). {We considered 7 metrics, in addition to accuracy. Details and results are given below}.\\

\noindent\textbf{Metrics.}
The metric {should} focus on several {points}. 
The first one is the error/success on predicting if the DNN model has some knowledge about specific data.  This involves detecting if the data is OOD or not. {For that,} we use three solutions proposed {in}~\cite{hendrycks2016baseline}. We first only {used} the confidence score of the OOD data and {on} the in distribution test data. Based on these confidence scores,  {and as in}~\cite{hendrycks2016baseline,hendrycks2019anomalyseg}, we evaluated the AUC, AUPR and the FPR-95\%-TPR, {that are indicators of the accuracy of detecting OOD data} 

{However}, these measures {give no information about the number of good predictions (that should be high) and of bad predictions (that should be low).} 
{This information is crucial since, although it is important to have a low score with the OOD data, the DNN should also reach a high confidence score for well-classified data,  and low confidence scores elsewhere. In case the DNN does not reach this point then it might be unusable.}

For that, the authors in \cite{corbiere2019addressing} propose to use metrics similar to the one used by Hendrycks \emph{et al.} ~\cite{hendrycks2016baseline} but rather than classifying into classes ``OOD'' or ``In distribution'', they classify as ``correctly classified'' or ``not correctly classified'' (this latter class contains both bad predictions and predictions on OOD data, see  \cite{corbiere2019addressing} for more details).

We also used the Expected Calibration Error (ECE) \cite{guo2017calibration}, which uses the $M$-bin histograms of confidence scores and accuracy.
The ECE performs a bin-to-bin difference between the two histograms, than an average over the $M$ bins. \Gianni{Similarly to \cite{guo2017calibration} we set $M =15$.}
{This metric, by measuring the difference between the expected accuracy and confidence, is an indicator of the quality of the confidence, and should be close to 0.}

\noindent\textbf{OOD classification  with MNIST \cite{lecun1998gradient}.} 
Concerning the classification, we used in a first experiment MNIST \cite{lecun1998gradient} which is a dataset composed of digit images as training dataset and NotMnist ~\cite{NotMnist} which {contains} letter images as OOD dataset. We first trained a classifier to learn to recognize the images of digits then tested it on the test set of MNIST and NotMnist hoping that the classifier would distinguish {digits form letters}. The DNN used for this experiment is fully connected and composed of 3 layers as in \cite{lakshminarayanan2017simple,franchi2019tradi}. Results are shown in Tables  \ref{table:OODresults1} and \ref{table:OODresults2} (MNIST rows).\\

\begin{table}[t!]
\renewcommand{\figurename}{Table}
\renewcommand{\captionlabelfont}{\bf}
\renewcommand{\captionfont}{\small} 
\begin{center}
\caption{Comparative results obtained on the Calibration task. }\label{table:OODresults1}
\resizebox{0.7\columnwidth}{!}{
\begin{tabular}{c l r r r r r }
\toprule
Dataset  & OOD technique  & Accuracy/mIoU & AUC~~ & AUPR & AUPR & ECE (\%) \\ 
         &               &  &  & Error & Success & \\

\midrule
    \multirow{6}{*}{\shortstack[c]{\ubold{MNIST/Not MNIST} \\ 3 hidden layers}}   & Baseline (MCP) &     98.8     &  92.7    &      96.1      &   81.4           & 0.305 \\
    & MCP + One class SVM       &     98.8    &   97.4  &       98.4     &      95.9 &   0.072   \\ 
    & MC Dropout      &     98.2     &   88.1  &       89.8     &      81.7 &   0.494   \\ 
         & Deep Ensemble  &  \ubold{98.9}   &  97.7   &  98.4  & 95.8        &      0.462      \\  
 &   TRADI &     98.6     &    97.1   &  98.4        &    94.6   & 0.407  \\  
                              & ODIN           &    98.8       &  94.2   &       96.8      &      85.6       &  0.500   \\ 
                               & ConfidNET      & 98.2         &   97.4  &      98.8      &   94.1           &  0.461      \\ 
                                & Ensemble OVA (ours)   &   97.2   & \ubold{99.0}   &    \ubold{99.5}   &  \ubold{97.3}        &          0.179   \\ 
                                & OVNNI (ours)   &   98.8   & \ubold{99.1}   &    \ubold{99.6}   &  \ubold{97.9}        &          0.066   \\ 
\midrule
                
\multirow{6}{*}{\shortstack[c]{\ubold{CIFAR10} \\ResNet50}}                                 & Baseline (MCP) &      93.1        &  83.9   &     92.9     &      67.5     &  0.606        \\  
                              & MCP +One class SVM  &   93.1        &  79.7   &     90.9     &      63.5     &  0.203              \\ 
                               & MC Dropout  &   93.1        &  83.9   &     92.9     &      67.5     &  0.606              \\ 
                               & Deep Ensemble  &  \ubold{95.0}   & \ubold{95.8}&   \ubold{97.7}  &    \ubold{92.1}  &  0.422\\ 
         &  ODIN           &  93.1&  83.9  &    93.3     & 67.2       &  0.606           \\ 
                                & ConfidNET      &  93.1& 85.1  &  94.6    & 61.2    &     0.706          \\
                                & Ensemble OVA (ours)   &  89.3  &  91.8 &   95.8  &   87.1 &     0.468         \\ 
                              & OVNNI (ours)   &  93.3  &  94.3 &   \ubold{97.3}  &   \ubold{91.1} &     \ubold{0.187}         \\ 
\midrule
\multirow{6}{*}{\shortstack[c]{\ubold{Camvid} \\ ENET}}                                & Baseline (MCP) &  85.8/52.9 & 79.7 &   52.1   &   92.6     &  0.146   \\ 
                                & MC Dropout      & 80.3/48.6  &     80.2          & 56.1    &       89.3         &      0.168       \\ 
                               & Deep Ensemble  &   88.0/58.2 &   83.2  &   54.3          &   94.0          &    0.112          \\  
              & TRADI          &    83.4/51.4  &   83.2 &  55.9   & 93.8           &    0.110          \\ 
                                & ConfidNET      &  83.4/52.8        & 81.3   &    58.3        &       92.6       &  0.121     \\ 
                                & Ensemble OVA (ours)       &  87.9/52.8        & 91.7  &    69.6        &       98.4       &  0.060    \\ 
                                 & OVNNI (ours)   & 93.1/66.1  & \ubold{94.0}    & \ubold{75.7}     &  \ubold{99.0}     & 0.025     \\ 
\midrule
\multirow{6}{*}{\shortstack[c]{\ubold{StreetHazards} \\ PSPNet (ResNet50)}}                               & Baseline (MCP) &        90.0/54.6 &   91.6   &  50.8     &   98.9    & 0.055 \\ 
                                 & MC Dropout      &   88.0/47.9       & 88.8    & 51.8      &  97.8     &   0.092   \\  
                                & Deep Ensemble  &    90.2/55.0      & 92.2    &  52.0      &  99.0       & 0.051  \\ 
            & TRADI          &    90.2 /54.6   &   92.1    &   51.4       &   99.1    &  0.049     \\ 
                                  & ConfidNET      & 90.0/54.6  & 88.9   &  37.0&     97.9      &  0.10            \\
                                & Ensemble OVA  (ours)   & 89.7/54.0      &  92.4   &   52.3     &     \ubold{99.1}         &  \ubold{0.048}  \\ 
                               & OVNNI (ours)   & 90.0/54.6    &  \ubold{93.0}   &   \ubold{53.4}     &     \ubold{99.2}         &  \ubold{0.048}  \\ 
\midrule
     \multirow{6}{*}{\shortstack[c]{\ubold{BDD Anomaly} \\ PSPNet (ResNet50)}}                           & Baseline (MCP) &    89.9/52.8      &  81.4   &  62.5   &     91.5    &     0.159         \\  
                               & MC Dropout      & 88.7/49.5 &  76.0   &       55.7     &     88.2         &    0.181          \\ 
                                & Deep Ensemble  &  91.0/57.6 &     85.5    &     67.3      &          93.9     &  0.170            \\ 
             & TRADI          &    89.9/52.1     & 81.9  &  63.2        &      91.8       &    0.157    \\ 
                  & ConfidNET      &    89.9/52.8      & 78.3    &   56.4         &   91.2           &   0.232     \\ 
                    & Ensemble OVA (ours)   &    89.9/52.8    & \ubold{91.2}     &    86.2    &     95.7         &      \ubold{0.072}         \\
                                & OVNNI (ours)   &    90.7/55.4    & \ubold{91.9}     &    \ubold{86.6}    &     \ubold{95.9}         &      \ubold{0.081}         \\ 
\bottomrule
\end{tabular}
} 
\end{center}
\vspace{-0.5cm}
\end{table}

\begin{table}[t!]
\renewcommand{\figurename}{Table}
\renewcommand{\captionlabelfont}{\bf}
\renewcommand{\captionfont}{\small} 
\begin{center}
\caption{Comparative results obtained on the OOD task. }\label{table:OODresults2}
\resizebox{0.6\columnwidth}{!}
{
\begin{tabular}{ c l r r r }
\toprule
Dataset   & OOD technique  & AUC  & AUPR & FPR-95\%-TPR \\ \midrule
   \multirow{6}{*}{\shortstack[c]{\ubold{MNIST/Not MNIST} \\ 3 hidden layers}}
                 & Baseline (MCP) & 94.0 & 96.0 & 24.6         \\  
                 &   MCP + One class SVM   & 96.9 & 98.0 & 12.5       \\ 
                  & MC Dropout      & 91.8 & 94.9 & 35.6         \\  
                  & Deep Ensemble  & 97.2 & 98.0 & 9.2          \\  
                & TRADI          & 96.7 & 97.6 & 11.0         \\  
               & ODIN           & 94.9 & 96.7 & 17.5         \\  
                 & ConfidNET      & 97.9 & 99.0 & 12.7         \\  
                 & Ensemble OVA (ours)   & 98.9 &\ubold{ 99.4} & 5.9         \\
                & OVNNI (ours)   & \ubold{99.3} &\ubold{ 99.6} & \ubold{3.5}         \\
                
\midrule

\multirow{6}{*}{\shortstack[c]{\ubold{CIFAR10} \\ResNet50}}                   & Baseline (MCP) & 80.4 & 89.7 & 61.5       \\  
                 &   MCP + One class SVM   & 78.8 & 89.6 & 61.5       \\  
                 & MC Dropout      & 80.4 & 89.7 & 62.6       \\  
                & Deep Ensemble  & \ubold{93.0} & \ubold{96.2} & \ubold{19.3}         \\  
                  & ODIN           & 80.3 & 89.9 & 61.3        \\  
                & ConfidNET      & 84.8 & 94.0  & 68.3        \\  
                  & Ensemble OVA (ours)   & 88.5 & 93.0 & 30.9        \\ 
                  & OVNNI (ours)   & 92.2 & \ubold{95.8} & 23.3        \\ 
\midrule
\multirow{6}{*}{\shortstack[c]{\ubold{Camvid} \\ ENET}}       & Baseline (MCP) & 75.4 & 10.0 & 65.1        \\  
                  & MC Dropout      & 75.4 & 10.7 & 63.2         \\  
                 & Deep Ensemble  & 79.7 & 13.0 & 55.3         \\  
                & TRADI          & 79.3 & 12.8 & 57.7         \\  
                  & ConfidNET      & 81.9 & 13.8 & 55.8         \\ 
                   & Ensemble OVA (ours)   & \ubold{97.1} & \ubold{71.1} & \ubold{13.5}         \\ 
                 & OVNNI (ours)   & \ubold{96.1} & \ubold{61.2} & \ubold{16.5}         \\ 
\midrule
                
  \multirow{6}{*}{\shortstack[c]{\ubold{StreetHazards} \\ PSPNet (ResNet50)}}              & Baseline (MCP) & 88.7 & 6.9  & 26.9         \\  
                  & MC Dropout      & 69.9 & 6.0  & 32.0         \\  
               & Deep Ensemble  & 90.0 & 7.2  & 25.4         \\  
       & TRADI          & 89.2 & 7.2  & 25.3         \\  
                   & ConfidNET      & 83.6 & 2.3   & 26.2            \\  
                   &  Ensemble OVA  (ours)   & \ubold{91.6} & \ubold{12.7} & \ubold{21.9}         \\ 
                & OVNNI (ours)   & \ubold{91.2} & \ubold{12.6} & \ubold{22.2}         \\ 
\midrule
\multirow{6}{*}{\shortstack[c]{\ubold{BDD Anomaly} \\ PSPNet (ResNet50)}}                 & Baseline (MCP) & 86.0 & 5.4  & 27.7         \\  
                 & MC Dropout      & 85.2 & 5.0  & 29.3         \\  
               & Deep Ensemble  & 87.0 & 6.0  & 25.0         \\  
                  & TRADI          & 86.1 & 5.6  & 26.9         \\  
                & ConfidNET      & 85.4 & 5.1  & 29.1         \\ 
                 & Ensemble OVA  (ours)   & 87.0 & 4.9  & 29.0         \\ 
                  & OVNNI (ours)   & \ubold{87.2} & \ubold{6.7}  & \ubold{25.0}         \\ 
\bottomrule
\end{tabular}
} 
\end{center}
\vspace{-0.5cm}
\end{table}

\noindent\textbf{OOD classification with CIFAR10~\cite{krizhevsky2009learning}.}
W also trained a network on CIFAR10 composed of classes airplanes, cars, birds, cats, deer, dogs, frogs, horses, ships and trucks. We have considered as OOD SVHN dataset \cite{Netzer2011}. Many papers \cite{maddox2019simple} train on CIFAR10 and test on the test set of CIFAR10 with noise or on STL-10 \cite{coates2011analysis}. It turns out that the first test aims more at measuring random uncertainty and the second one the {capability to adapt to the domain}. {We rather have preferred} to consider as an OOD  dataset SVHN which is a color image dataset of digits, {that guarantees} that the OOD data really comes from a distribution different from that of CIFAR10. The DNN we used on this experiment is Resnet50  \cite{he2016deep}, which has the advantage of being popular in the community.
Results are shown in Tables \ref{table:OODresults1} and \ref{table:OODresults2} (CIFAR10 rows).

\noindent\textbf{OOD Segmentation with Camvid~\cite{brostow2008segmentation}.} We used Camvid, a dataset conventionally used in {works dealing with} segmentation or uncertainty theory and deep learning ~\cite{kendall2015bayesian,corbiere2019addressing,franchi2019tradi}. {This is dataset is an ``easy'' dataset but allows you to quickly validate results}. 
To test the ability of OVNNI to detect OOD pixels, we trained on all Camvid classes except 3 classes (pedestrian, bicycle, and car), that we deleted, by marking the corresponding pixels as unlabeled. These three classes correspond to  OOD classes. 
Thus this experimental protocol proposed by ~\cite{franchi2019tradi} makes it possible to validate that the trained DNN will detect the pixels on which it has not been trained as OOD. The DNN for this experiment is Enet~\cite{paszke2016enet}. Results are shown in Tables \ref{table:OODresults1} and \ref{table:OODresults2} (Camvid rows).\\
\noindent\textbf{OOD Segmentation with StreetHazards~\cite{hendrycks2019anomalyseg}.}
StreetHazards is a large-scale dataset that contains {different sets} of synthetic images of street scenes. More precisely, this dataset is composed of 5125 images for training and 1500 test images.
The training dataset {contains} 13 classes and the test dataset is composed of the 13 training classes and 250 OOD classes, making it possible to test the robustness of the algorithms with all possible scenarios. 
For this experiment we used PSPnet~\cite{zhao2017pyramid} with the experimental protocol in~\cite{hendrycks2019anomalyseg}. The architecture used for the PSPnet is ResNet50.
Results are shown in Tables \ref{table:OODresults1} and \ref{table:OODresults2} (StreetHazards rows).

\noindent\textbf{OOD Segmentation with BDD Anomaly~\cite{hendrycks2019anomalyseg}.}
BDD Anomaly dataset is a subset of BDD dataset, composed of 6688 street scenes for the training set and 361 for the testing set. The training set {contains} 17 classes, and the test dataset is composed of the 17 training classes and 2 OOD classes.
For this experiment we used PSPnet~\cite{zhao2017pyramid} with the experimental protocol in~\cite{hendrycks2019anomalyseg}]. The architecture used for the PSPnet is ResNet50.
Results are shown in Tables \ref{table:OODresults1} and \ref{table:OODresults2}  (BDD Anomaly rows).


\subsection{Discussions}

On MNIST we can see in Tables \ref{table:OODresults1} and \ref{table:OODresults2} that OVNNI has competitive results for detecting OOD data; more specifically, its calibration score (ECE) is the best. With respect to the metrics proposed by Hendryck \emph{et al.}, OVNNI is the most effective in detecting OOD images, improving the best AUC by 1.4\% the best AUPR by 0.6\% and the best FP of 63.2\%. Concerning the metrics proposed by Corbi\`{e}re \emph{et al.}, OVNNI  improves the AUC by 1.41\%, the AUPR Error by 0.80\% the AUPR success by 2.14\% and the ECE by 70.6\%. 

\begin{figure}[!t]
     \centering
         \subfloat[][]{\includegraphics[width=0.25\linewidth]{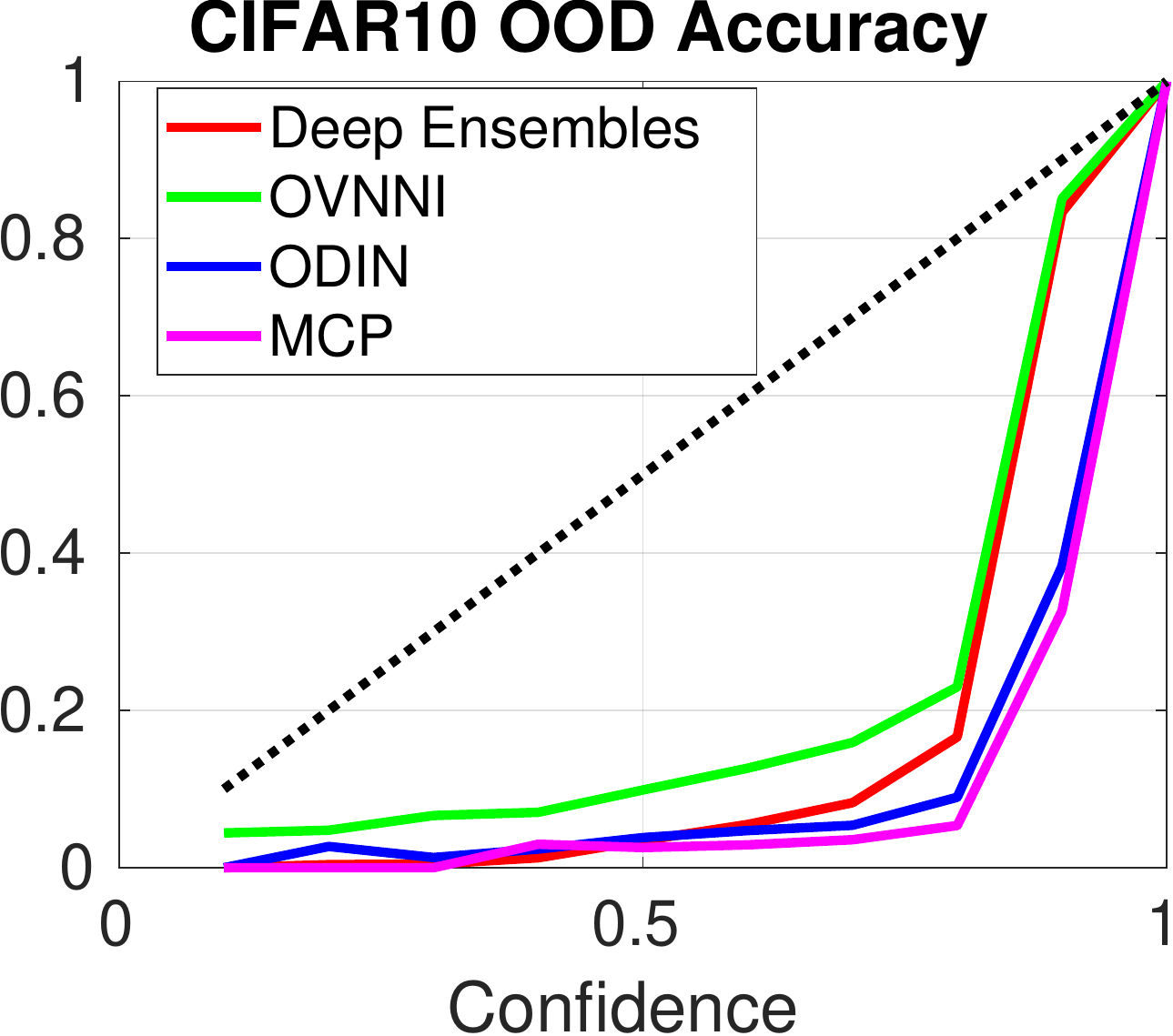}\label{cifar10}}
         \subfloat[][]{\includegraphics[width=0.27\linewidth]{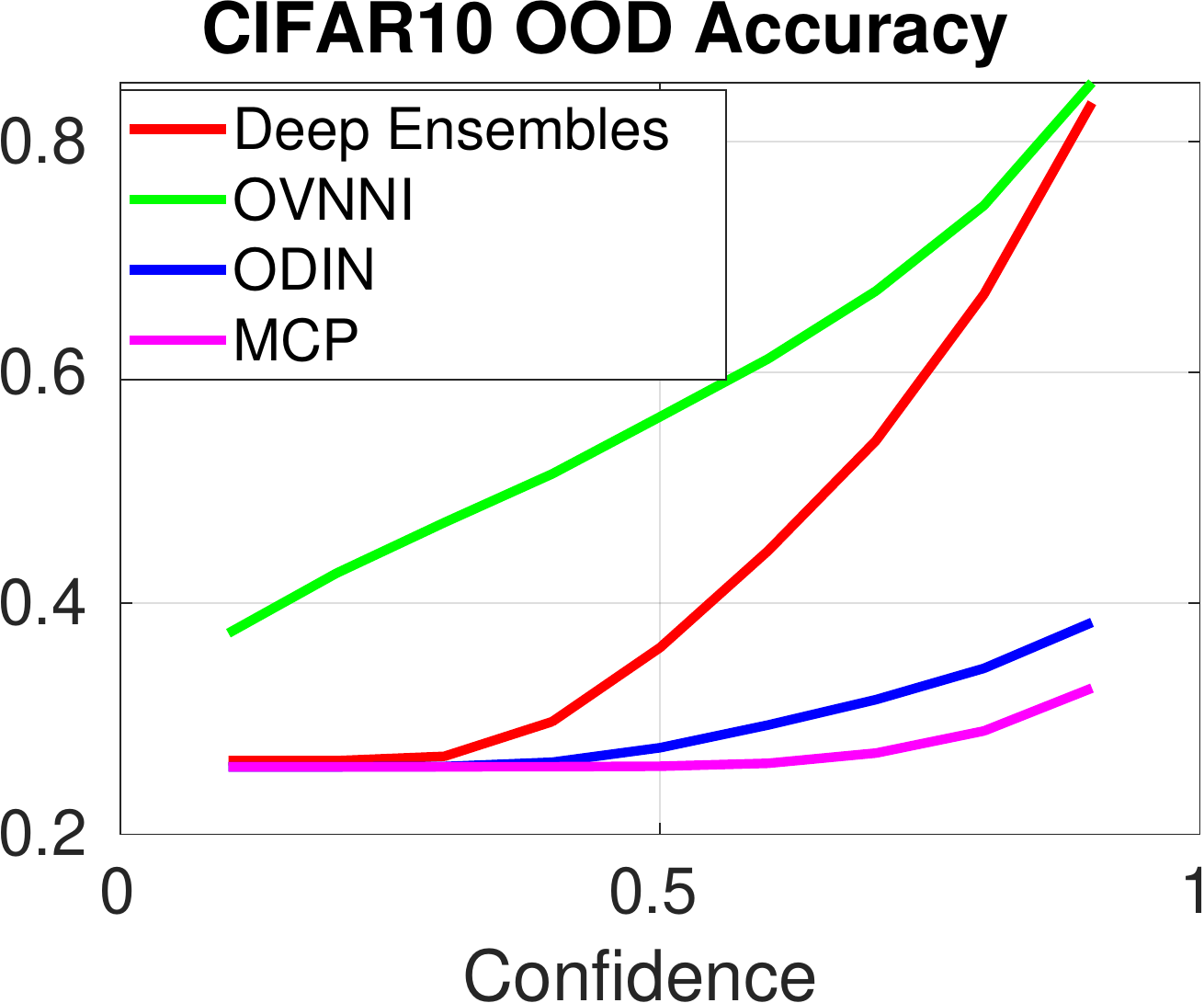}\label{cifar10_calib}}
         \subfloat[][]{\includegraphics[width=0.23\linewidth]{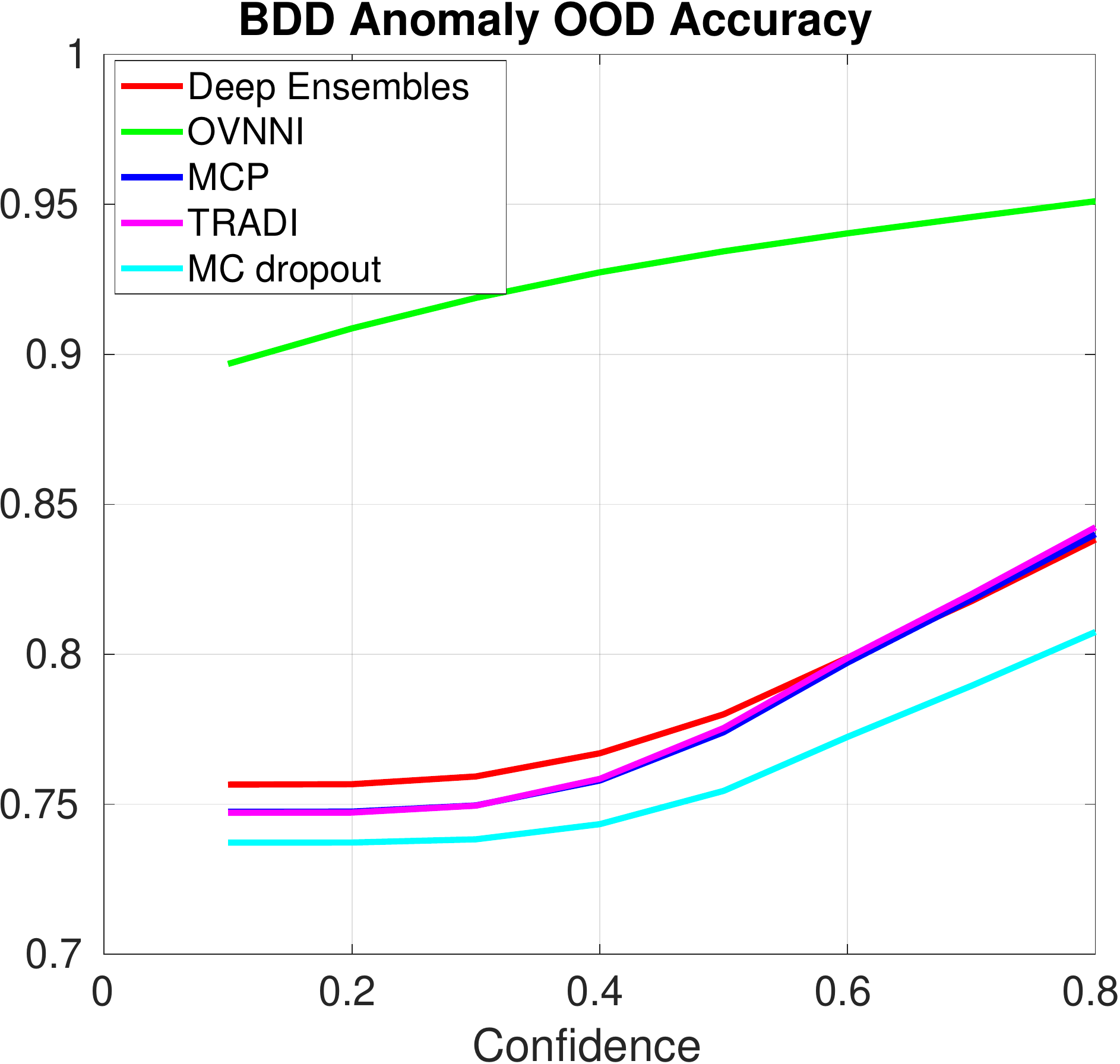}\label{BDD}}
         \caption{(a) and (c) Accuracy vs confidence plot on the CIFAR10 \textbackslash SVHN and BDD Anomaly experiments, respectively.  (b)  calibrationn plot on the CIFAR10 \textbackslash SVHN.}
     \label{fig:all}
         \end{figure}

On CIFAR10, 
although Deep Ensembles achieve good results on all the measurements as well except on the ECE, note that OVNNI is better calibrated. This can also be seen in the histogram in Figure \ref{fig:all}. The difference between OVNNI and Deep Ensembles is low and the crucial requirement of DNN is to have a good calibration. Hence, having a good calibration is more important than having a good AUC or AUPR. Also, we have represented the \textit{accuracy vs confidence} curves in Figure \ref{fig:feature_space}. These curves are defined in \cite{lakshminarayanan2017simple} and are constructed by evaluating the accuracy of all data where the DNN has reached confidence thresholds.
These curves show the performance of the OVNNI confidence index over CIFAR10. Finally, we have illustrated the OVNNI calibration on CIFAR10 in the calibration curve in Figure \ref{fig:feature_space}.  The calibration plot is defined on \cite{guo2017calibration} and is constructed by taking bins of data based on their confidence score. Then on each bin, we evaluate the accuracy, as it should ideally be comparable to the confidence score. These curves show once again the good performance of OVNNI in terms of calibration.

On Camvid we note that OVNNI improves the results of the state of the art by up to 77\% with regard to the metrics proposed by Corbi\`{e}re \emph{et al.}~\cite{corbiere2019addressing}, and by up to 77\% for calibration as well. Concerning the metrics proposed by Hendryck \emph{et al.},  OVNNI improves the measurements by a maximum of 22\%. 

On StreetHazards we show in Table \ref{table:OODresults2} that OVNNI has better results than the state of the art by improving the best results by up to 42.8\%. In Table \ref{table:OODresults1} OVNNI improves the result by a least 2.6\% and improves state-of-the-art ECE by 2\%. These results show the interest of using OVNNI for semantic segmentation.

Finally, on BDD Anomaly OVNNI improves the calibration by at least 48\% which is highly relevant, given the importance of this metric. Furthermore concerning the other metrics, OVNNI improves the results by at least 22\%. Furthermore, in Figure \ref{fig:all} we have illustrated the confidence accuracy curve of several algorithms. These curves  underline again that OVNNI reaches the best performance in terms of calibration.

Overall, these results show that OVNNI improves the calibration of networks by rendering the confidence in their results more in line with their expected results. Making DNN models more reliable is crucial, especially in areas where the model should not be overconfident. In \cite{guo2017calibration} the authors show that good accuracy of DNNs comes with a price, namely their reliability. In this work, we propose a solution that increases accuracy in most cases, while at the same time improving the calibration and the OOD detection performance. The conceptual simplicity of this solution is a significant asset for its adoption, and the results also convey the message that 'one vs all training' can still have an interest for a finer understanding of epistemic uncertainty in DNNs.

 \begin{figure}[!thb]
 \begin{center}
 \begin{tabular}{c c c c c}
 \includegraphics[width=0.20\linewidth]{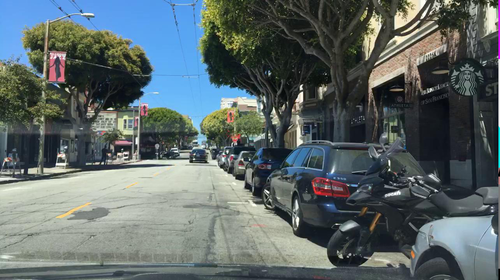}&
 \includegraphics[width=0.20\linewidth]{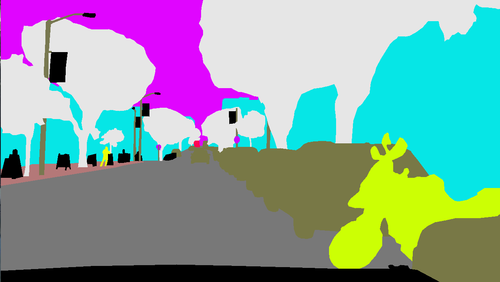}&
  \includegraphics[width=0.20\linewidth]{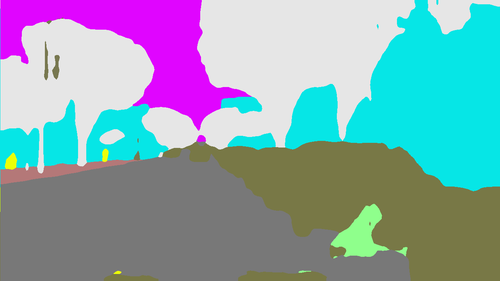}&
    \includegraphics[width=0.20\linewidth]{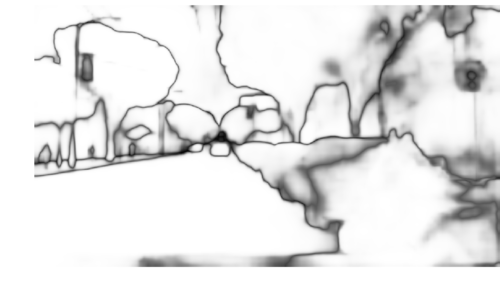}
   \includegraphics[width=0.20\linewidth]{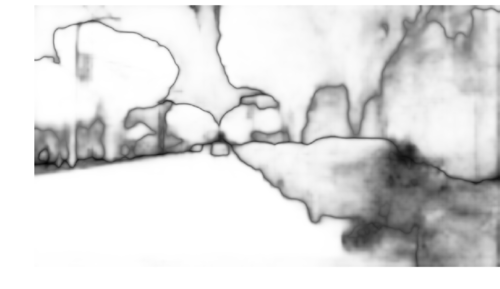}\\
 \includegraphics[width=0.20\linewidth]{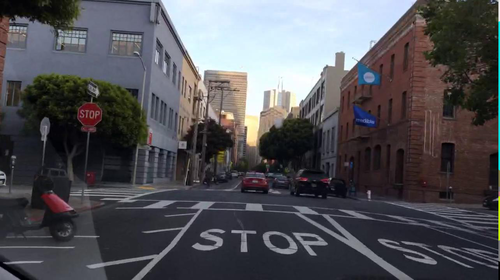}&
 \includegraphics[width=0.20\linewidth]{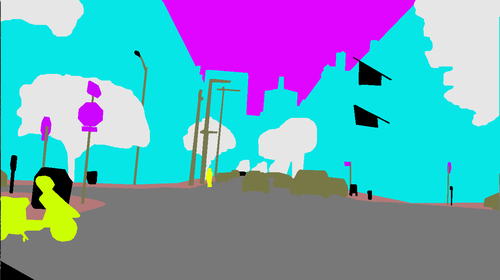}&
  \includegraphics[width=0.20\linewidth]{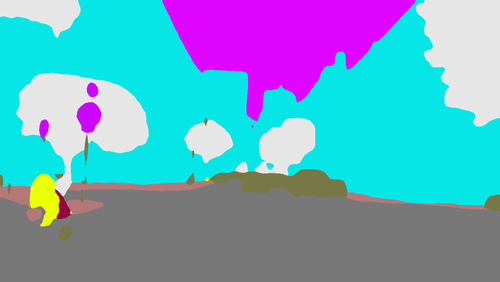}&
    \includegraphics[width=0.20\linewidth]{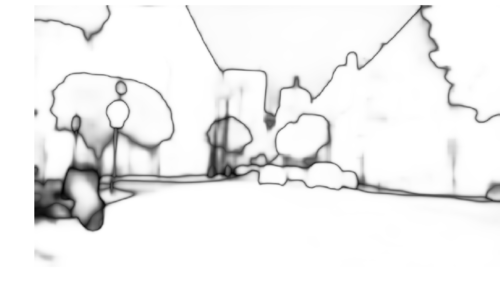}
   \includegraphics[width=0.20\linewidth]{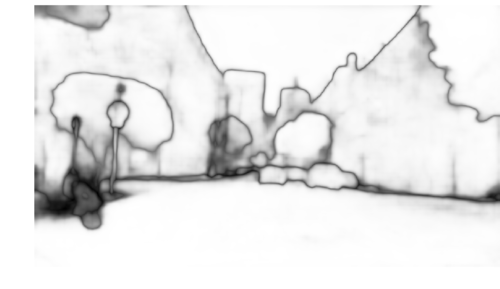}\\
          \includegraphics[width=0.20\linewidth]{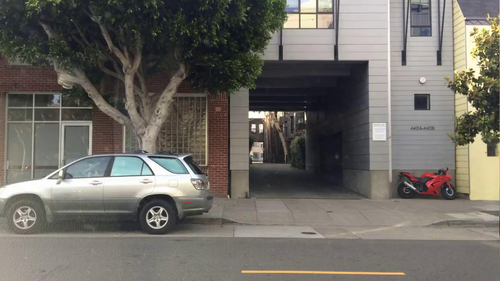}&
 \includegraphics[width=0.20\linewidth]{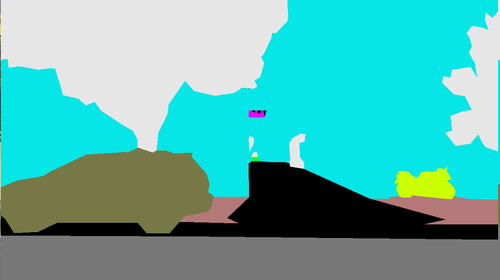}&
  \includegraphics[width=0.20\linewidth]{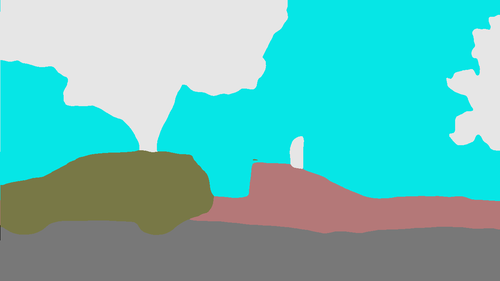}&
    \includegraphics[width=0.20\linewidth]{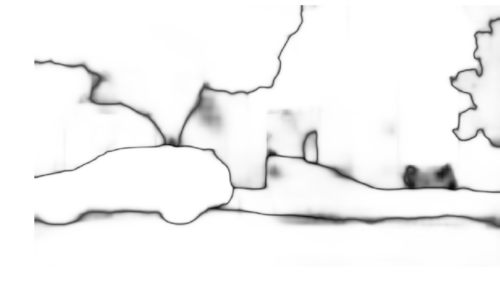}
   \includegraphics[width=0.20\linewidth]{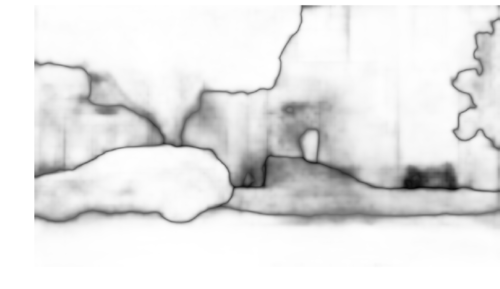}\\
    \includegraphics[width=0.20\linewidth]{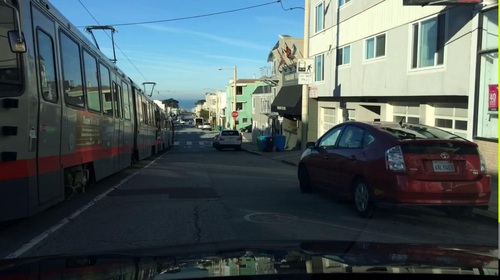}&
 \includegraphics[width=0.20\linewidth]{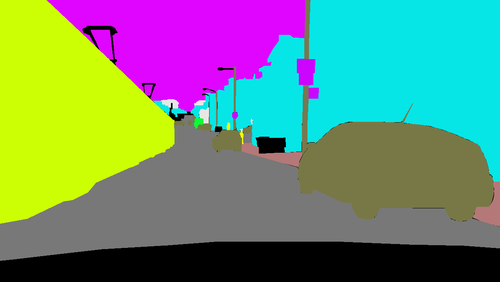}&
  \includegraphics[width=0.20\linewidth]{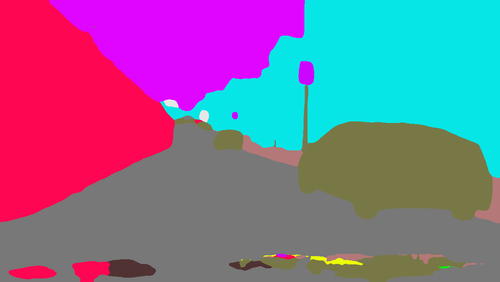}&
     \includegraphics[width=0.20\linewidth]{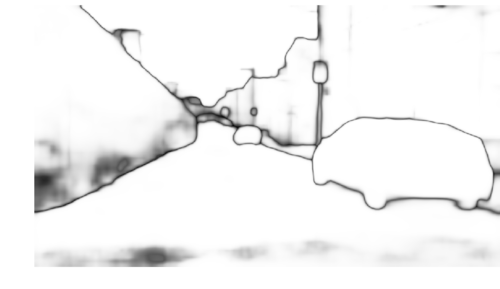}
   \includegraphics[width=0.20\linewidth]{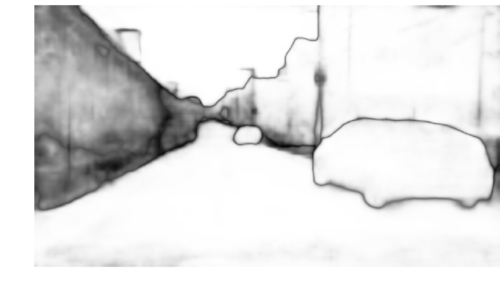}
 \end{tabular}
 \end{center}
 \caption{Results of OVNNI on BDD Anomaly. The first column is the input image, the second is the ground truth, the third is prediction and the fifth is the confidence score of OVNNI. \Gianni{For comparison, we add the MCP confidence score in the fourth column.} We can see that OVNNI has a low score on the motorcycle on the three first rows and on the train on the last row which correspond to the OOD classes.}
  \label{fig:accuconfidance}
   \end{figure}

 \begin{figure}[!thb]
 \begin{center}
 \begin{tabular}{c c c c c}
 \includegraphics[width=0.20\linewidth]{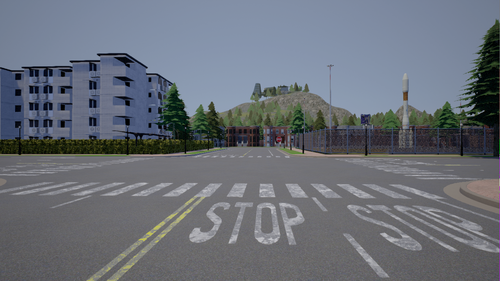}&
 \includegraphics[width=0.20\linewidth]{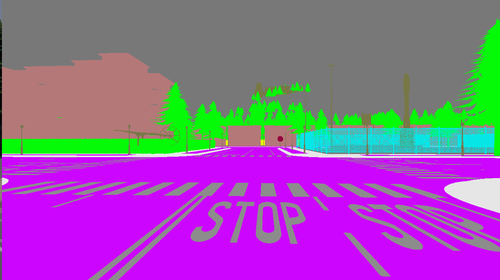}&
  \includegraphics[width=0.20\linewidth]{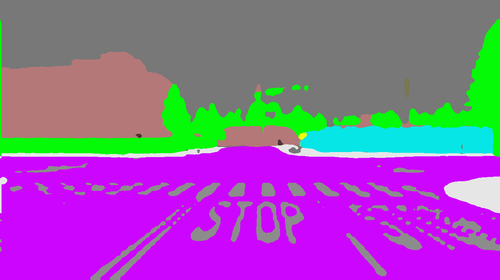}&
      \includegraphics[width=0.20\linewidth]{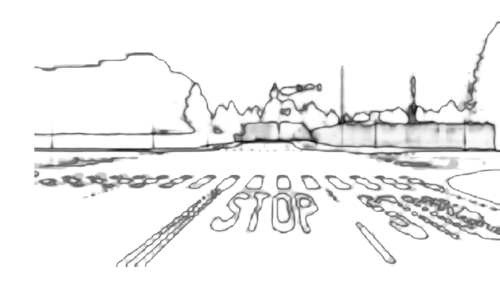}&
   \includegraphics[width=0.20\linewidth]{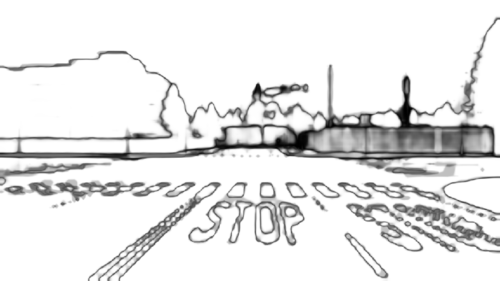}\\
 \includegraphics[width=0.20\linewidth]{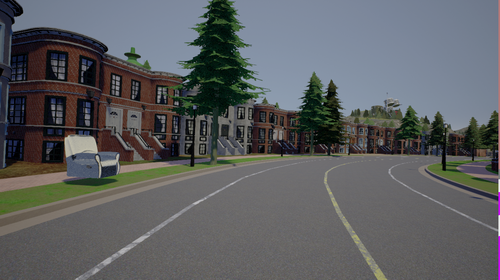}&
 \includegraphics[width=0.20\linewidth]{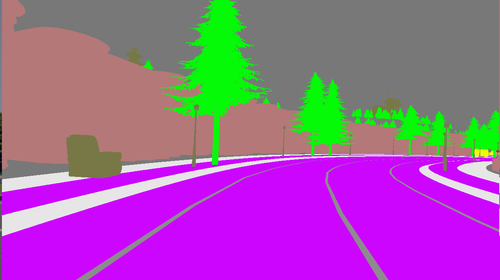}&
  \includegraphics[width=0.20\linewidth]{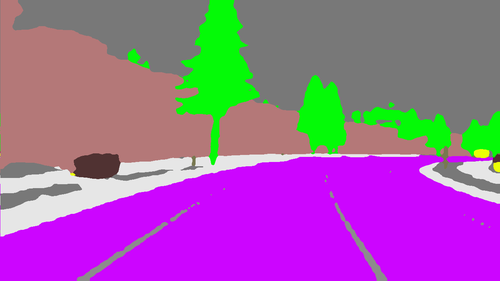}&
      \includegraphics[width=0.20\linewidth]{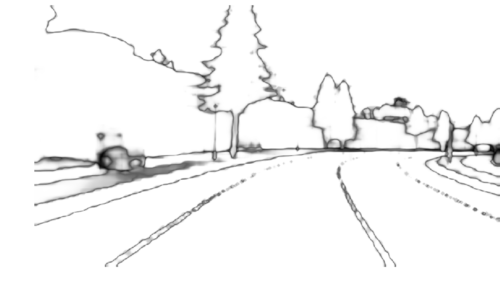}&
   \includegraphics[width=0.20\linewidth]{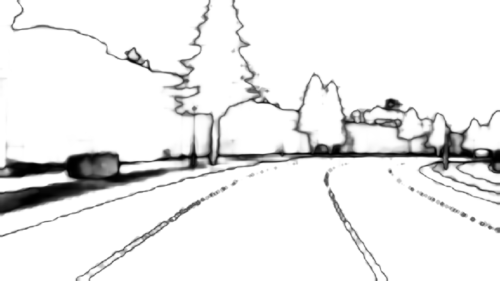}\\
          \includegraphics[width=0.20\linewidth]{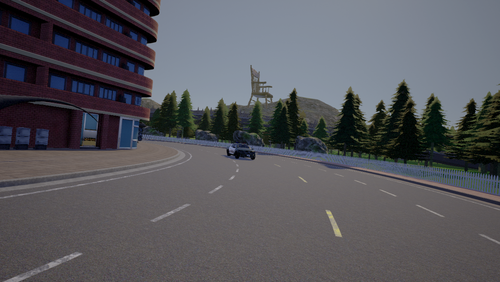}&
 \includegraphics[width=0.20\linewidth]{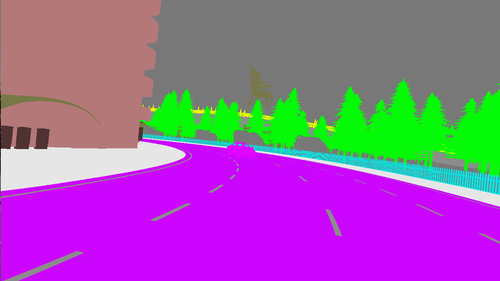}&
  \includegraphics[width=0.20\linewidth]{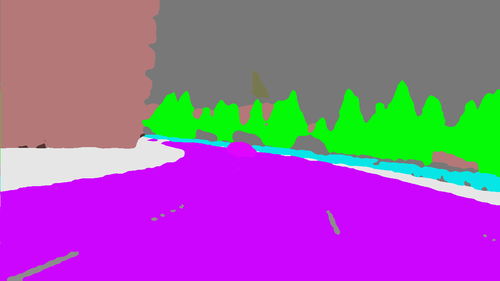}&
      \includegraphics[width=0.20\linewidth]{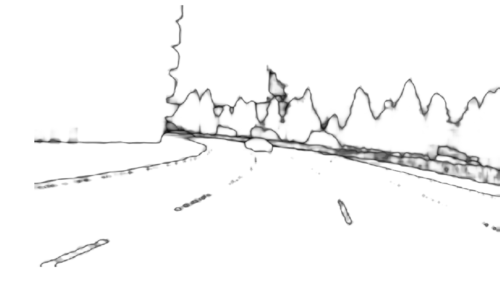}&
   \includegraphics[width=0.20\linewidth]{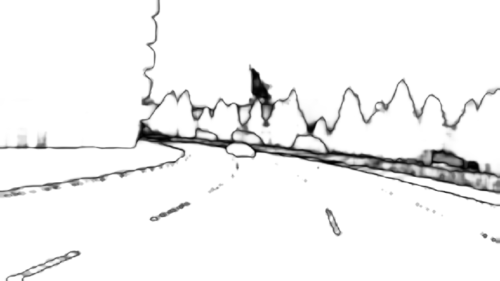}\\
    \includegraphics[width=0.20\linewidth]{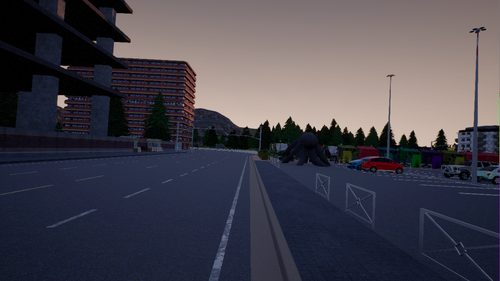}&
 \includegraphics[width=0.20\linewidth]{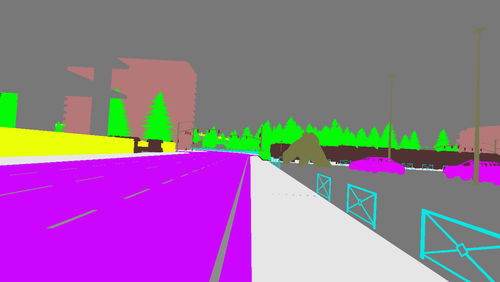}&
  \includegraphics[width=0.20\linewidth]{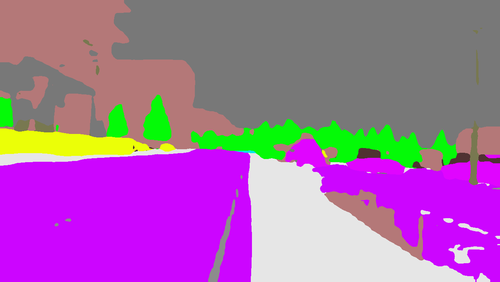}&
    \includegraphics[width=0.20\linewidth]{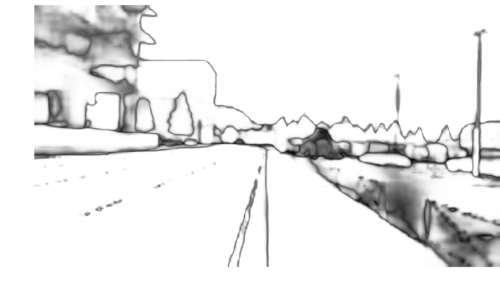}&
   \includegraphics[width=0.20\linewidth]{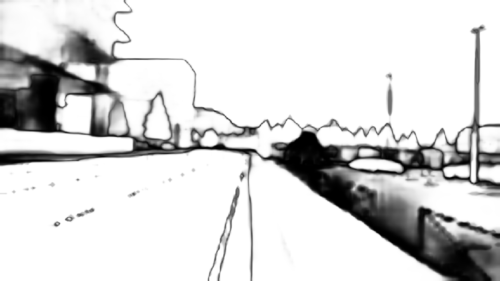}
 \end{tabular}
 \end{center}
 \caption{Results of OVNNI on StreetHazards. The first column is the input image, the second is the ground truth, the third is prediction and the last is the confidence score of OVNNI. \Gianni{For comparison, we add the MCP confidence score in the fourth column.} We can see that OVNNI has a low score on the chair, the seat, the rocket and the spider which correspond to the OOD classes.}
  \label{fig:accuconfidance}
   \end{figure}

\section{Conclusions}
In this work, we presented an approach based on One versus all training and mixed with a modern approach based on deep learning. We show that the combination of these approaches reaches states of the art performance on all segmentation experiments. Regarding classification tasks, OVNNI exhibits the best calibration performance. Concurrent approaches suffer from a lack of performance in calibration in most datasets, hence the scores that they provide are overconfident, potentially leading to dangerous scenarios in critical applications. In addition to the reported performance, our approach needs little hyperparameter tuning and is easy to implement.  

Future work involves first extending this strategy to new tasks such as medical image analysis.  One could also use this framework for active learning since active learning algorithms require techniques that can detect OOD data. 

\clearpage
%
%
\bibliographystyle{splncs04}
\bibliography{egbib}
\end{document}

%% file: related.tex
 \section{Related work}
 
 OOD detection is not a novel problem and has been studied before the deep learning revival in various branches of machine learning under slightly different taks: anomaly~\cite{liu2008isolation}, outlier~\cite{breunig2000lof} or novelty detection\cite{scholkopf2000support}. In the last few years, this task has seen increased attention from different communities and has been addressed with: \textit{predictive uncertainty estimation}, \textit{ensemble methods}, \textit{image reconstruction}, etc. In the following we review briefly some of the methods related to our approach.


\noindent\textbf{Classification with a \textit{background} class.}
In multiple computer vision tasks, \eg, object detection~\cite{ren2015faster,liu2016ssd}, it is common to use a \textit{background} class in addition to the known classes to classify. This leads to a better separation of the classification space and a more discriminative classifier. While this seems to be a reasonable and straightforward approach, for OOD detection, it is likely to suffer from negative dataset bias~\cite{tommasi2017deeper} and thus not generalize to other background objects not-seen during training. In our approach, we also use a part of the classes as background when training the individual classifiers, however the overlap of their decision boundaries coupled with the AVA model, better distinguishes in- from out-of-distributions samples.

\noindent\textbf{Anomaly detection by reconstruction.} Anomalies can be detected by training an autoencoder~\cite{creusot2015real,baur2018deep} or generative model~\cite{schlegl2017unsupervised,lis2019detecting} on in-distribution data and use the quality of the reconstruction as a proxy OOD as the autoencoder is unlikely to decode accurately patterns not seen during training. Training such models for accurate and robust reconstruction requires large amounts of data.

\noindent\textbf{Bayesian approaches} Bayesian Neural Networks (BNN)~\cite{neal1996} are elegant, intuitive and easy to reason models, that can capture the epistemic uncertainty through the exploitation of the distributions of their weights. In spite of recent progress that make them more tractable~\cite{blundell2015weight}, they are still limited to small or medium-size networks, while most DNNs in usually enclose millions of parameters.
Gal and Ghahramani~\cite{gal2016dropout} aimed for a method to imitate BNNs. To this end they proposed Monte Carlo Dropout (MC Dropout) to estimate the posterior
predictive network distribution by sampling different subsets of neurons at each forward pass during test time and aggregate their predictions. 
In computer vision, MC Dropout is the most popular instance of BNNs due to its speed and simplicity. It has been extended to other tasks, \eg, semantic segmentation~\cite{kendall2015bayesian}, pose estimation~\cite{kendall2015modelling}. 
However, the benefits of Dropout are more limited for convolutional layers, where specific architectural design choices must be made~\cite{kendall2015bayesian,mukhoti2018eval}. Recent OOD benchmarks for semantic segmentation~\cite{hendrycks2019anomalyseg,lis2019detecting} show that MC Dropout still admits many false positives.

\noindent\textbf{Ensembles.} Ensemble methods are prominent techniques for measuring epistemic uncertainty. They have the potential to encapsulate a true diversity in the weights of the composing models, contrarily to the dispersion introduced by MC Dropout~\cite{fort2019deep}, which ultimately focuses on a single mode. Lakshminarayan~\etal~\cite{lakshminarayanan2017simple} propose training an ensemble of DNNs with different initialization seeds. Vyas \etal~\cite{vyas2018out} train an ensemble of classifiers in a self-supervised way on different subsets of the training data, using the left-out data as OOD.
Izmailov \etal~\cite{izmailov2018averaging} collect weight checkpoints from local minima and average them or fit a distribution over them and sample networks~\cite{maddox2019simple}. Franchi \etal~\cite{franchi2019tradi} track weights trajectories across training and compute their distributions, further used for sampling an ensemble of networks. Our approach also exploits ensembles, however each network is specialized on a different classification task. We exploit the complementarity in this ensemble for better OOD predictions.  

 \noindent\textbf{Traditional use of OVA/OVO ensembles} These aggregation techniques are popular for performing multi-label classification based on an ensemble of binary base classifiers. For OVO, instead of the baseline max-voting aggregation strategy, pairwise coupling~\cite{wu2004probability} or ECOC~\cite{dietterich1994solving} have been widely used, but the quadratically increasing number of base classifiers may limit significantly OVO applicability in the case of large label sets. In contrast, OVA fusion uses a linearly increasing number of base classifiers, and relies in most works on a Winner-Takes-All class assignment based on the maximum class response. To the best of our knowledge, these ensembling methods have not been used for estimating  the epistemic uncertainty of DNNs.

\noindent\textbf{Deep OOD detection.} A recent line of approaches addresses OOD detection through DNNs specific heuristics. Hendrycks and Gimpel~\cite{hendrycks2016baseline} established a standard baseline for OOD detection relying on the Maximum Class Probability from softmax. \cite{devries2018} attach a confidence branch to a classification network and train it to predict OOD samples, while ODIN~\cite{liang2017enhancing} learns a temperature scaling for softmax values and adversarial perturbation to better distinguish OOD data. \Gianni{Lee \etal~\cite{lee2018simple} get a class conditional Gaussian distributions with respect to features that they tune on a dataset with OOD data and In distribution data. Lambert \etal~\cite{lambertmseg}  attenuate uncertainty by training on a large composite dataset leading to a more robust DNN. Zendel \etal~\cite{zendel2018wilddash} propose a semantic segmentation dataset for checking the confidence score of DNNs. The authors of \cite{bevandic2019simultaneous}  train a DNN to predict OOD confidence score.}
Lee \etal~\cite{lee2017training} train a GAN along with the classifier to produce near-distribution examples and enforce lower classifier confidence on GAN samples. Malinin and Gales~\cite{malinin2018predictive} use Dirichlet networks to build a distribution over the prediction distributions for OOD detection. Most of these methods rely on a OOD dataset during training and are likely to specialize on specific anomalies from this data ~\cite{hendrycks2018deep}. In contrast, in our approach we do not require OOD examples during training, as we leverage the multiple one-versus-all classifiers.

 